# On Stable Multi-Agent Behavior in Face of Uncertainty (Preliminary Report)

**Moshe Tennenholtz**
Faculty of Industrial Engineering and Management
Technion–Israel Institute of Technology
Haifa 32000, Israel

## Abstract

A stable joint plan should guarantee the achievement of a designer's goal in a multi-agent environment, while ensuring that deviations from the prescribed plan would be detected. We present a computational framework where stable joint plans can be studied, as well as several basic results about the representation, verification and synthesis of stable joint plans.

## 1 Introduction

*Stable behavior* is central to decision-making in multi-agent environments. There are several forms of stability which have been discussed in the literature. In some cases stability refers to rationality in multi-agent encounters. A stable rational behavior (aka. Nash-equilibrium (Fudenberg & Tirole 1991)) is a joint behavior of the agents, such that it is irrational for each of the agents to deviate from it, assuming the other agents stick to their part of that behavior. Other types of stability aim to optimize the behavior of one party (e.g., a designer, the government (Kreps 1990)) assuming the other parties are rational. In addition, work in control-theory (Ramadge & Wonham 1989), and work in distributed systems (Dwork & Moses 1990), have faced a different, although somewhat related, types of stability, where we wish a system to behave in a robust manner in face of machine failures.

A major theme of the above-mentioned lines of research is the interest in systems where deviations from a prescribed behavior are either irrational, impossible, or do not affect the overall activity of the system. Unfortunately, in many situations this stability can not be obtained. For example, it might be unreasonable to assume that goals can be obtained given arbitrary types of machine failures.

In this paper we discuss a basic multi-agent setting, where the designer's objective is to devise a joint plan for a group of agents in order to obtain a desired goal. The basic objective is that the goal will be obtained whenever the agents follow the prescribed plan. However, a major additional requirement is that if an agent deviates from the joint plan then this deviation should be detected by the other agents. This is especially important if this deviation leads to a situation where the designer's goal is not obtained. Hence, assuming we have a pair of agents, each of which might observe only part of the other agent's states and actions, and given that the designer's goal is composed of particular states to be reached by both agents, the designer's aim would be to devise a *stable joint plan*. In a stable joint plan, a deviation by one of the agents will be detected by the other agent. The designer can then be informed about that deviation.

One can easily observe that the above-mentioned form of stability is useful for many settings. In the case where agents might deviate from their plans due to machine failures, stability may enable the designer to detect that the system operates in an unsatisfactory manner. In the case of rational agents, where an agent might maliciously deviate from a suggested joint plan, stability may prevent deviations by enabling the use of punishment mechanisms (for which the detection of deviations is crucial) (Brafman & Tennenholtz 1996).

In most of this paper our interpretation to stability is in the flavor of mechanism design in Game Theory (Fudenberg & Tirole 1991; Myerson 1991). In a mechanism design problem a designer, who has her own preferences and desires, has to devise an appropriate behavior (e.g., joint plan) to a group of agents. However, the agents might deviate from prescribed behaviors that are irrational for them. Rationality (and, respectively, irrationality) is captured in game theory by the concept of equilibria. In an equilibrium, a deviation by an agent is irrational since it will not increase its payoff/utility, given that the other agents would stick to their prescribed behavior. The related Game-Theoretic notions assume that the utility functions of the agents are common-knowledge (or, in more elaborated settings, that there exists a probability distribution on the utility functions which is common knowledge). This is a point that we wish to relax in this paper. In order to handle this we use the following idea. Each agent is required to execute a particular plan and to notify the designer about any deviation from prescribed plans that it observes. Assuming the designer has the power to "punish" a deviating agent, no deviations that are to be exposed are rational. Hence, our notion of equilibria will be a qualitative one. The aim of the designer is to devise a plan that will guarantee that her goal will be obtained and that will be stable against any possible deviation: any deviation from a prescribed plan which will prevent the achievement of the designer's goal will lead the deviating agent to a low payoff since that deviation will be detected by the other agents and will lead to a proper punishment.

In order to illustrate the above-mentioned problem and approach consider a pair of robots, controlled by different programmers, which are about to work in a shared environment. The designer may require the robots to perform a particular task, and may wish to ensure that this task is actually performed. She will then devise a plan for performing that task that is stable against deviations by any single agent. This plan will have the property that any deviation by an agent which will prevent goal achievement will be detected by the other agent. If robot 1 (i.e., its programmer) decides to use a track that is different from the track prescribed to it by the designer's plan and this deviation prevents the achievement of the designer's goal, stability will ensure that this deviation will be detected and announced by robot 2, which will cause an appropriate reaction to the designer.

In the following sections we present a computational setting where stable joint plans can be studied, as well as several results about the representation, verification, and synthesis of stable joint plans. We start with a basic computational setting, where uncertainty is a result of potential deviations from prescribed behaviors. In a later point, we extend the discussion to the case where there is incomplete information about the environment behavior, in addition to the potential deviations by the agents. Most of our computational study refers to moderate settings, where the number of states in the system is taken to be polynomial in the actual representation size. This is done in order to concentrate on the issue of stability, rather than tackle the general intractability of planning in (even simple)

non-moderate settings. [1]

## 2    The Computational Setting

### Definition 2.1:

Let $S = \langle S_1, S_2, B, A_1, A_2, T \rangle$ be a *system*. A system consists of a pair of *agents*,[2] where agent $i$ is associated with a set of states $S_i$, and with a set of actions $A_i$, and $B$ is a set of *environment states*. $T : S_1 \times S_2 \times B \times A_1 \times A_2 \to S_1 \times S_2 \times B$ is a *state transition function*. Let $C = S_1 \times S_2 \times B$ denote the set of *system configurations*.

A system consists of a pair of agents, each of which has its own states and actions. At each point, each agent can observe only its state. In addition, there is another component to the system, that we refer to as the system environment, which has its own state. The system is initialized in an *initial configuration* $c_0 \in C$. A *goal* for the system is a set of configurations $C_g \subseteq C$. A *plan* for agent $i$ is a sequence of actions.[3] Given an initial configuration $c_0$ and a goal $C_g$, we say that the *joint plan* $P = (P_1, P_2)$ (where $P_i$ is the plan for agent $i$) is *efficient* if $P$ leads from $c_0$ to an element $c_g \in C_g$ and no more than $|C|$ actions are executed in the course of $P_i$ $(i = 1, 2)$.

A goal is obtained when a configuration $c_g \in C_g$ is reached. Notice that an agent can not in general deduce from its state that a goal configuration has been reached. This point is a major motivation for our study. If the achievement of a goal could be deduced from the state of a single agent then this agent could easily detect deviations by the other agent which prevent goal achievement. Unfortunately, this is usually not the case. A goal can be a composite goal which requires each of the agents to reach a particular state. For example, one agent may be required to reach a location $l_1$ while the other agent may be required to reach a location $l_2$; the fact that the first agent has reached $l_1$ need not necessarily guarantee that the second agent has reached $l_2$. Similarly, the agents may need to perform certain actions which will bring the environment to a particular state. However, when reaching an appropriate goal state, the state of an agent might not contain enough information to guarantee that the goal has been actually obtained.

---

[1] See however Section 7 where crash failures in non-moderate settings are discussed.

[2] Our discussion and results can be easily extended to the case where the number of agents is a constant $c$, where $c \geq 2$.

[3] In a later point, when we discuss incomplete information about the environment behavior, we will be interested in a more complicated type of plans.

The model we have defined is a general model for multi-agent environments. Consider the robots mentioned in the introduction. In order to be more concrete assume that the robots (i.e., agents) move in a grid-like environment such as those found in assembly plants or in a warehouse (see (Shoham & Tennenholtz 1995) for a more detailed discussion). At each point in time each robot can either stay in its place or move to one of its neighboring coordinates. When a robot is in a particular coordinate, it can see every coordinate which is in a distance of one coordinate from it. In general, rows and columns of the grid correspond to lanes, while the rest of space is devoted to equipment and material the robots may need to use. The designer may require the robots to move along particular paths and carry different items from place to place. The state of each robot represents its location and observations in a particular point, and the environment state can refer to additional information such as the location of particular tools. Notice that the knowledge of an agent about its environment need not be explicitly represented as part of its state. An agent's state of knowledge is generated based on its history of observations and actions, and its background knowledge about the system and the plans to be executed by the agents. The actions taken by the agent may change the agents' state and the environment state. As the reader can see our model can represent in a quite straightforward manner situations as the one mentioned above. The semantics of our model follows the general semantics for distributed systems discussed in (Fagin *et al.* 1995), and its more concrete details and their use are discussed in (Moses & Tennenholtz 1995; Safra & Tennenholtz 1994).

Our basic assumption is that the structure of $\mathcal{S}$, as well as the identity of $c_0$ and $C_g$, are common-knowledge. This point will be relaxed in Section 6. Once a joint plan has been prescribed to the agents, it becomes common-knowledge as well. However, after the system has been initialized, each agent can observe only its states. Naturally, an agent may deduce, based on its observations and initial knowledge, some facts about the other agent's state or actions.

It is easy to see that if there is a joint plan leading from the initial configuration to a goal configuration, then there is also an efficient joint plan which obtains the desired goal. In fact, the computation of an efficient joint plan is a simple graph search (Aho, Hopcroft, & Ullman 1974). This refers to the case where the agents stick to the prescribed behavior. In our setting, however, an agent might deviate from that behavior. This leads to the definition of stable joint plans.

**Definition 2.2:**

Given a system $\mathcal{S}$, an initial configuration $c_0$, and a goal $C_g$, a joint plan $P = (P_1, P_2)$ for the agents is *stable* if the following hold:

1. $P$ is efficient.

2. If agent $i$ deviates from $P_i$ while the other agent ($j$) sticks to $P_j$, and $C_g$ is not reached (by the corresponding joint plan), then this deviation will be detected[4] by $j$ (for $i = 1, 2$, and $j \neq i$).

In the following sections we study several computational aspects of stable joint plans. In particular, we will consider the verification of stability, as well as the synthesis of stable joint plans.

## 3 The Verification of Stability

The verification that a joint plan achieves a desired goal is easy. However, in order to have a stable joint plan, we need to ensure that deviations which prevent goal achievement would be detected.

Given a joint plan, we wish to verify whether agent 1 can detect deviations by agent 2 (while agent 1 follows its plan). The case where agent 2 detects deviations by agent 1 is treated similarly.

We now present a polynomial algorithm for the verification that a deviation by agent 2, which prevents goal achievement, will be detected by agent 1. The input of the algorithm is a system and an efficient joint plan $P = (P_1, P_2)$ leading from the initial configuration of the system to a goal configuration. Notice that, in general, the agents need not halt when arriving at a goal configuration $c_g$. In the sequel, we assume that the number of configurations and actions is polynomial in the actual representation size.

Let $t \leq |C|$ be the length of $P$ (i.e., the total number of actions executed in the course of $P$). Let $C' = C \times N$, where $N$ is a set of time stamps; the range of $N$ is between 0 and $t$.

**The Detection Algorithm:**

1. Mark as "good" all elements of $C'$ which correspond to time $t$, in which the state of agent 1 is as prescribed by the joint plan (i.e., as it should be by the end of $P$'s execution).

---

[4] We require the detection of the fact a deviation has occurred. We do not require a detection of the exact nature of this deviation.

2. Mark as "good" each element $c'_1 = (c_1, k) \in C'$, for which the following conditions hold:

- The state of agent 1 in $c_1$ is as the state it should be in when the time is $k$ (when both agents conform to the joint plan).

- The joint plan specifies an action $a \in A_1$, such that when augmented with some action $b \in A_2$ (which may be the action prescribed by the plan, but might be an action which deviates from the plan) will lead from $c_1$ to an element $c_2$ where $c'_2 = (c_2, k+1)$ is already marked as "good".

3. If $(c_0, 0)$ is marked as "good" and a path of length $t$ which passes only through "good" elements but not through $C_g$ exists, then return failure (a deviation is possible). Otherwise, go back to 2. If there is a round in which no new element has been marked as "good", then return success (no deviation is possible).

**Theorem 3.1:**

*The detection algorithm is polynomial, and announces success if and only if no deviation is possible.*

## 4 Eventual Detection of Deviations

In the previous section we have shown a polynomial algorithm for the verification that a given joint plan is stable. In the setting studied in the previous section, the number of configurations is taken to be polynomial in the actual representation size, and agents have complete information about the environment behavior. All the uncertainty in the system is a result of deviations made by the agents. These facts, augmented with the positive result obtained in the previous section, suggest one may wish to consider the feasibility of an automatic synthesis of stable joint plans.

**Definition 4.1 :** [The stable joint plan problem](SJPP): Given a system $\mathcal{S}$, an initial configuration $c_0$, and a goal $C_g$, find a stable joint plan, if exists, and otherwise announce that no such plan exists.

As the following theorem shows, the automatic synthesis of stable joint plans is intractable. The reduction used in the proof of the following theorem sheds light on the structure of stable joint plans. It shows that deviations might lead to situations where the designer might need to solve constraint satisfaction problems in order to enable one agent to distinguish between deviations and proper behavior of the other agent.

**Theorem 4.2:**

*The SJPP is NP-complete.*

**Proof:** (Sketch):

We prove the theorem by reduction from 3-SAT. Let $\varphi$ be a 3-CNF formula, where $C_1, C_2, \ldots, C_m$ are the clauses in $\varphi$. The set of primitive propositions is $X = \{x_1, x_2, \ldots, x_n\}$. For every $i$ $(1 \leq i \leq m)$, clause $i$ consists of three literals $l_{i_1}, l_{i_2}$ and $l_{i_3}$, where each $l_{i_j}$ is a primitive proposition or its negation.

We take the environment to have a unique state (hence, this state is fixed and cannot be changed, and we need not refer to it in the transition function). The goal $C_g$ will contain only one element. We have two agents. Agent 1 has $n + 4$ states: $s_0, s_1, \ldots, s_n, s_{n+1}, p, q$, where $s_0$ is the initial state of agent 1, and $s_1$ is the part of the goal associated with agent 1. Agent 2 has $m + 3$ states: $r_0, r_g, r_1, r_2, \ldots, r_m, o$, where $r_0$ is the initial state of agent 2, and $r_g$ is the part of the goal associated with agent 2.

Agent 1 can perform the following actions. In the initial state it can perform only the action $a$, while in $s_i (1 \leq i \leq n)$ it can perform only the actions $x_i$ and $\neg x_i$, which are associated with the corresponding literals. In $s_{n+1}$ agent 1 can perform only the action *observe*. In $p$ and $q$ the agent can not perform any action (or, alternatively, it can perform only the null action).

Agent 2 can perform the following actions. In the initial state it can perform actions $d, d_1, d_2, \ldots, d_m$. In all other states agent 2 can perform only the null (no) action.

It remains to define the transition function. The joint action $(a, d)$ will lead from the initial configuration to the goal configuration, and $a$ will always lead from $s_0$ to $s_1$. Each action of the form $x_i$ or $\neg x_i$ will lead from the corresponding $s_i$ to $s_{i+1}$. The effects of *observe* in $s_{n+1}$ will be as follows: *observe* will lead to $p$ if and only if agent 2 is not in $o$; if agent 2 is in $o$ then the state $q$ will be reached.

By performing $d_i$ agent 2 will move from $r_0$ to $r_i$. Any action which corresponds to a literal in the $j$'s clause will move agent 2 from $r_j$ to $o$.

Now, if there exists a satisfying assignment, then by choosing the plan which corresponds to it as the plan for the first agent, and requesting agent 2 to perform $d$ as its first action, we will guarantee the desired stable plan. This is due to the fact that we will reach the state $o$ in a case of a deviation by agent 2 (and regardless of the nature of that deviation). On the other hand, if there is a stable plan then we must ensure that agent 2 will reach either its goal state or the state $o$. Otherwise, agent 1 will not be able to distinguish between deviations and goal achievements. In order to do so, agent 1 must choose a satisfying literal for each of the clauses. In addition, it can not choose a literal and its negation by the construction (of agent 1). This yields the desired result. ∎

## 5   Fast Detection of Deviations

The negative result obtained in the previous section suggests one may wish to consider modifications and restrictions of our basic setting, which may make the synthesis problem tractable. One interesting modification of the synthesis problem is discussed below.

Stable joint plans should enable the detection of deviations. In many realistic systems, however, eventual detection of deviations is not satisfactory. In particular, deviations might cause various kinds of damage; the faster these deviations are detected, the better the system would work. For example, in our robotics setting, when a robot takes a track which is different from the one prescribed by the designer, it may lead to conflicts with other tasks that are under the designer's control (e.g., the maintenance of a particular machine); hence, a fast detection of the corresponding deviation might become crucial. This suggests that deviations should be detected as fast as possible, and leads to the following definition.

### Definition 5.1:

Let $k$ be an integer. Given a system $\mathcal{S}$, an initial configuration $c_0$, and a goal $C_g$, a joint plan $P = (P_1, P_2)$ is $k$-stable if the following hold:

1. $P$ is an efficient joint plan leading from $c_o$ to $C_g$.

2. A deviation from $P_i$ ($i = 1, 2$) at time $m$, will be detected by agent $j \neq i$ no later than at time $m + k + 1$.

In order to handle $k$-stable plans we will use the following construction. Let $C' = C \times T$, where $T$ denotes the non-negative integers between 0 and $|C|$. Let $P(c')$ denote the set of available joint plans in configuration $c'$ [5], which are of length $k + 1$. Define $\bar{C} = \{(t, c', p(c')) : t \in T, c' \in C, p(c') \in P(c')\}$. Let $c_1 = (t_1, c', p(c'))$, and $c_2 = (t_2, c'', p(c''))$. We say that $c_2 \in \bar{C}$ is reachable from $c_1 \in \bar{C}$ if the following hold:

1. $t_2 = t_1 + 1$.

2. The first joint action in $p(c')$ leads from $c'$ to $c''$.

3. The $k$ last actions in $p(c')$ coincide with the $k$ first actions in $p(c'')$.

4. A deviation by one of the agents from its part of the first joint action,[6] will be detected while the other agent performs its $k$ following actions.

Notice that the number of elements in $\bar{C}$ is polynomial in $|C|$, and that points (1),(2) and (3) are easy to check. Point (4) can be verified as follows. Let $c'' = (b, s_1, s_2)$, and for simplicity consider the case where agent 1 is the deviating agent. We enumerate all possible plans of length $k + 1$ of agent 1, which may be initialized in $c'$. If there exists such a plan, which when executed in parallel to the plan prescribed to agent 2 by $p(c')$, leads in the first step to $(b', s', s_2)$ (where $s' \neq s_1$ or $b' \neq b$), and on the following steps agent 2 will visit the same states as it would visit when the agents follow $p(c')$, then there exists a deviation which can not be detected. If no such plan has been found then every deviation will be detected. Notice that the verification of point (4) is polynomial.

We denote the set of pairs $(c_1, c_2) \in \bar{C}^2$, where $c_2$ is reachable from $c_1$, by $E(\bar{C})$.

Consider now the following stable joint plan algorithm [SJPA], which relies on the previous definition and algorithm for the generation of $\bar{C}$ and $E(\bar{C})$.

1. Denote by "good" all elements of $\bar{C}$, for which the first joint action specified in the corresponding joint plan (of length $k + 1$) leads to the goal, and any deviation by one of the agents from that joint action will be detected in the following $k$ steps (as in (4) above),

2. Search for an element $(c_1, c_2) \in E(\bar{C})$, where $c_2$ is denoted by "good" and $c_1$ is still not marked as "good". If such an element has been found, then denote $c_1$ by "good" and go back to 2.

3. If an element $c$ which is associated with the initial configuration has been marked as "good" then return "success"; otherwise, return "failure".

### Theorem 5.2:

*SJPA is polynomial, and generates a $k$-stable joint plan iff such a plan exists.*

---

[5] Recall that a plan for an agent in this case is simply a sequence of actions; there is no a-priori requirement that it should reach a particular state.

[6] Such a deviation should lead to a configuration which differs from the configuration obtained if both agents obey the prescribed joint plan.

## 6   Uncertainty and Incomplete Information

We have discussed stable joint plans in the context of settings where uncertainty is a result of (malicious or non-malicious) deviations from a prescribed behavior. Uncertainty about the environment state has been a direct consequence of the potential deviations. In this section we wish to relax this assumption. This is obtained by removing the assumption that the initial configuration $c_0$ is common-knowledge. In the rest of this paper we assume that there is a set $C_0 \subseteq C$ of *possible initial configurations*. The system may be initiated in any configuration $c_0 \in C_0$, but the identity of the actual initial configuration is initially unknown. We assume that the agents are able to initially communicate about their initial states, but the initial environment state is unknown and may be any $b_0 \in B$ which is consistent with the agents' knowledge about $C_0$ and their initial states (our discussion and results hold also for the case where the agents are not able to communicate about their initial states).

The above-mentioned setting is a general setting for planning with incomplete information. The idea of introducing incomplete information by presenting lack of knowledge about the initial configuration is standard in game theory. For example, in our robotics example, it might be a-priori unknown whether a particular movement turns on machine 1 or machine 2. This incomplete information may be expressed by the environment state. The effects of the different actions is a function of the environment state, and by having incomplete information about the initial environment state we can capture uncertainty about the effects of actions, as well as other types of uncertainty (e.g., uncertainty about the actual state of the environment).

Notice that although the state of the environment is initially unknown, the agents may learn about the state of the environment based on their observations. A *plan* for an agent operating in a setting with incomplete information is a decision tree where at each node a decision about the action to be performed is made based on the recent observation (i.e., state which has been reached). This is a standard description of conditional plans (Safra & Tennenholtz 1994). Notice that this representation makes the action-selection to depend on the agent's *history* of observations and actions. In particular, an agent $i$ may reach a state $s \in S_i$ twice during the course of its plan, select an action $a$ in its first arrival at $s$, and select an action $b \neq a$ in its second arrival at $s$. This is due to the fact the state $s$ refers to an observable state of the agent, rather than to its complete knowledge state which is built based on its history of observations and actions. We will say that a joint plan is efficient if it guarantees the achievement of the desired goal for any possible $c \in C_0$, while the number of actions which might be executed in a course of the plan is bounded by $|S_1| \times |S_2| \times |B|$. A major point to notice is that efficient plans in this context might, in principle, be exponentially large. This crucial point is discussed in detail below.

The reader should notice that although we assume we can list the set of possible environment states, the problem of coming up with a satisfactory plan might be quite problematic even when we have a single agent. This is due to the fact that a plan in this case may become a decision tree of exponential size. Indeed, as Safra and Tennenholtz have shown (Safra & Tennenholtz 1994) the problem of coming up with a plan which will guarantee goal achievement in the single-agent case is NP-hard. Fortunately, they have also shown that if there is a plan leading to the desired goal, then there is such a plan which can be represented in polynomial space and be verified in polynomial time.

We now extend the above-mentioned positive result to the case of a pair of agents which operate with incomplete information about the environment, where plans are required to be also stable.

**Theorem 6.1:**

*Given a system with incomplete information, if a stable joint plan exists, then there is such a plan that can be encoded in polynomial space and be verified in polynomial time.*

The proof of the above theorem is omitted from this abstract. It follows the proof by Safra and Tennenholtz for the single-agent case (Safra & Tennenholtz 1994). We wish to emphasize that this result, when augmented with the result presented in Section 4, shows the importance of plan-design processes for the construction of stable joint plans. More specifically, we have shown that the synthesis of stable joint plans is intractable even when there is no uncertainty about the environment state. In fact, the proof of Theorem 4.2 has taken the environment state to be fixed. However, an off-line construction of a stable joint plan by a trial and error procedure may become feasible, due to the fact that stable joint plans can be efficiently represented and verified. The latter holds even for settings where uncertainty about the environment state does exist (in addition to the potential agent deviations). Assume that in our robotics example the designer can enumerate the set of relevant environment states, although she may not be able to a-priori know what the actual initial configuration is. The designer may wish to equip the robots with a joint plan which will guarantee the achievement of her goal, and that will be stable. Our result supplies the designer with a guarantee that if an appropriate stable joint plan exists then it can be encoded efficiently. Moreover, when a candidate for an appropriate plan is suggested, the verification of whether it is satisfactory or where it fails can be carried out efficiently. This enables the use of a trial and error procedure for the generation of the desired plan.

## 7    Non-Moderate Settings

Until this point we have assumed that our systems are moderate. In a moderate system, the number of possible configurations is taken to be polynomial in the actual representation size (or, alternatively, we measure complexity as a function of the number of configurations). Notice that this restriction does not imply that the number of knowledge states that an agent may reach is small (see (Fagin *et al.* 1995; Safra & Tennenholtz 1994; Moses & Tennenholtz 1995)). In particular, the latter might be exponentially larger than the number of agent's states. Hence, moderate settings are still an appealing and non-trivial context.

Consider now a non-moderate setting of incomplete information. We still assume that the number of environment states is polynomial in the actual representation size, but we assume that the number of agents' states might be exponential in the actual representation size. For example, one may consider a situation where the state of an agent is described by a set of $n$ primitive propositions. In the latter case we may wish to consider representations which are polynomial in $n$ (and therefore the number of agents' states in this case is exponential in the actual representation size). In non-moderate settings it makes no sense to associate efficiency with the achievement of a goal in $|B| \times |S_1| \times |S_2|$ steps, since this value might be exponential in the actual representation size. Hence, we will be interested in plans where the number of actions which might be executed in a course of a plan is polynomial in the actual representation size.

When we have exponentially many states then even single agent planning with complete information and without any possible deviation becomes intractable. It is easy to check that this implies that the verification that a given joint plan is stable is intractable in this case. However, we now show a common class of non-moderate settings of incomplete information, where the representation and verification of stable joint plans can be carried out efficiently.

Crash failures are deviations in which an agent might suddenly fail, and stay in that situation from that point on (Dwork & Moses 1990; Moses & Tennenholtz 1995). When an agent crashes, it starts to repeatedly execute a distinguished action, "null". The effects of "null" on the system behavior will be determined by the transition function.

**Theorem 7.1:**  *Consider a non-moderate system with incomplete information, and assume that all possible deviations are crash failures. If a stable joint plan exists, then there is such a plan which can be encoded in polynomial space and be verified in polynomial time.*

## 8    Discussion

In this paper we have introduced a computational study of stable joint plans. The aim of stable joint plans is to enable goal achievement, while detecting deviations which might prevent the achievement of the desired goal. This type of stability is useful for a variety of multi-agent settings. In particular, it enables a designer to obtain a desired goal, while ensuring that if the goal has not been obtained due to a deviation by one of the agents, then she would be announced about that by the other agent. In the case of malicious failures, stability is a major tool in the prevention of deviations.

On the conceptual level, in this paper we have introduced a qualitative form of mechanism design. Work on multi-agent systems in AI has adopted the Game-Theoretic mechanism design approach (Rosenschein & Zlotkin 1994; Kraus & Wilkenfeld 1991; Sandholm & Lesser 1995). However, the related Game-Theory literature has made strong assumptions about the knowledge available to the agents. At first glance, it might be somewhat unclear whether useful mechanisms can be devised without such knowledge. Our work suggests a qualitative mechanism, which does no rely on classical assumptions on the agents' utility functions, but relies on the assumption that a designer has the power for punishing a deviating agent. As in classical mechanism design the designer wishes to obtain her goal, but in the lack of knowledge about the objectives of agents she would devise a joint plan that will lead to the detection of deviations. Given the ability to punish deviating agents, we get a qualitative stability concept in which the designer's goal is obtained and no deviations are rational.

On the more technical level our work extends results obtained for single-agent planning with incomplete information to the multi-agent case. Previous work has shown that off-line design processes are useful in the context of single-agent planning with incomplete information. In particular, single-agent planning with incomplete information is intractable in moderate settings, and it is off-line tractable (i.e., the corresponding plans can be efficiently represented and verified) even in quasi-moderate settings (i.e., when the number of environment states is polynomial but the number of agents' states might be exponentially large). Our work reveals similar phenomena in the multi-agent context. In this context, incomplete information is replaced or augmented with potential deviations by the agents. In this case we show that in moderate settings, the existence of potential deviations (while we have complete information about the environment) suffices to make the generation of stable plans intractable, while the related problem is off-line tractable even if the potential deviations are augmented with incomplete information about the environment state.

Most of our study refers to moderate settings where the number of configurations is polynomial in the actual representation size. While it not our aim to advocate moderate representations, we wish to make clear the distinction between these representations and exhaustive representations of the agents' state space. The state of an agent in our setting refers to the physical/observable state of an agent, and not to its more general mental/local/knowledge state. The knowledge state of the agent is built incrementally and implicitly as a function of the agent's observations and actions. Notice that the number of knowledge states is exponentially larger than the number of agent's states to be actually represented. This is in fact the situated automata idea (Rosenschein 1985). Notice that our complexity results refers to the number of an agent's observable states rather than to the exponentially larger number of its mental/local/knowledge states. Needless to say, it may be of interest to consider in future work more succinct representations, where the number of observable states is exponential in the actual representations size.

We see the mixture of the qualitative approach to design with the more classical approach to mechanism design as a promising research direction. In particular, some knowledge about the utility function of agents may be available or deduced during plan execution, and may be integrated into the corresponding algorithms and enable the generation of better plans. The mixture of various kinds of failures, e.g., crash failures and rational malicious failures, would also require non-trivial adjustments to the related setting. In addition, although stable joint plans prevent deviations by rational agents, they do not fix deviations which are non-maliciously caused by agents who may wish to conform to suggested plans. The need to fix such deviations may be a subject for further study. We hope to address these topics in future work.